\documentclass[runningheads,pagebackref,breaklinks,colorlinks]{llncs}

\usepackage[year=2024,ID=866]{eccv}

\usepackage{eccvabbrv}
\usepackage{graphicx}
\usepackage{booktabs}
\usepackage{hyperref}
\usepackage{multirow}
\usepackage{multicol}
\usepackage{float, booktabs, boldline, hhline}
\usepackage{wrapfig}
\usepackage{soul}
\usepackage{xcolor}
\usepackage{color, colortbl}
\usepackage{tabularx}
\usepackage{bbding}

\usepackage{amsmath,amsfonts,bm}

\def\eqref#1{equation~\ref{#1}}

\def\1{\bm{1}}

\def\vc{{\bm{c}}}
\def\vd{{\bm{d}}}

\def\vm{{\bm{m}}}
\def\vn{{\bm{n}}}
\def\vo{{\bm{o}}}
\def\vp{{\bm{p}}}

\def\vv{{\bm{v}}}

\def\vx{{\bm{x}}}

\def\mN{{\bm{N}}}

\def\mP{{\bm{P}}}

\def\mS{{\bm{S}}}
\def\mT{{\bm{T}}}

\def\mY{{\bm{Y}}}

\DeclareMathAlphabet{\mathsfit}{\encodingdefault}{\sfdefault}{m}{sl}
\SetMathAlphabet{\mathsfit}{bold}{\encodingdefault}{\sfdefault}{bx}{n}
\newcommand{\tens}[1]{\bm{\mathsfit{#1}}}

\def\tS{{\tens{S}}}

\usepackage{pifont}

\usepackage[accsupp]{axessibility}  

\usepackage{orcidlink}

\begin{document}

\title{CF-PRNet: Coarse-to-Fine Prototype Refining Network for Point Cloud Completion and Reconstruction} 

\titlerunning{CF-PRNet}

\author{Zhi Chen\orcidlink{0000-0002-9385-144X} \and
Tianqi Wei \orcidlink{0009-0005-0134-6438} \and
Zecheng Zhao  \and
Jia Syuen Lim  \and
Yadan Luo \orcidlink{0000-0001-6272-2971} \and
Hu Zhang \orcidlink{0009-0009-9892-9515} \and
Xin Yu \orcidlink{0000-0002-0269-5649} \and
Scott Chapman \orcidlink{0000-0003-4732-8452} \and
Zi Huang \orcidlink{0000-0002-9738-4949}
}

\authorrunning{Z.~Chen et al.}

\institute{The University of Queensland, Australia \\
\email{\{zhi.chen,helen.huang\}@uq.edu.au}}

\maketitle

\begin{abstract}

In modern agriculture, precise monitoring of plants and fruits is crucial for tasks such as high-throughput phenotyping and automated harvesting. This paper addresses the challenge of reconstructing accurate 3D shapes of fruits from partial views, which is common in agricultural settings. We introduce CF-PRNet, a coarse-to-fine prototype refining network, leverages high-resolution 3D data during the training phase but requires only a single RGB-D image for real-time inference. Our approach begins by extracting the incomplete point cloud data that constructed from a partial view of a fruit with a series of convolutional blocks. The extracted features inform the generation of scaling vectors that refine two sequentially constructed 3D mesh prototypes—one coarse and one fine-grained. This progressive refinement facilitates the detailed completion of the final point clouds, achieving detailed and accurate reconstructions. CF-PRNet demonstrates excellent performance metrics with a Chamfer Distance of 3.78, an F1 Score of 66.76\%, a Precision of 56.56\%, and a Recall of 85.31\%, and win the first place in the Shape Completion and Reconstruction of Sweet Peppers Challenge \footnote{https://cvppa2024.github.io/challenges/}. Our source code is available at \url{https://github.com/uqzhichen/CF-PRNet/}.

\end{abstract}

\section{Introduction}
\label{sec:intro}

As the global population continues to surge, the agricultural sector faces the critical challenge of meeting an escalating demand for food. This situation is compounded by several factors, including climate change, a shortage of labor, and declining biodiversity. One promising solution to these challenges is the use of autonomous robotic systems, which can enhance agricultural productivity throughout the entire plant growth cycle—from sowing and fertilizing to irrigating and harvesting. Recent advances in artificial intelligence have spurred significant improvements in various agricultural tasks, including irrigation planning~\cite{gao2021ant}, plant disease recognition~\cite{wei2024benchmarking, wei2024snap,plantseg}, nutrient deficiency identification~\cite{zhang2023divide}, and fruit harvesting~\cite{onishi2019automated}.

This paper addresses the specific challenge of modeling the complete 3D shape of a sweet pepper from only partial observations. Unlike general object completion tasks that may benefit from diversity, fruit shape completion requires accurate reconstruction to reflect true fruit morphology, which is heavily influenced by environmental factors. The variability in potential fruit shapes, especially in greenhouse settings, presents a unique challenge due to \textbf{data scarcity} and significant \textbf{domain shifts}. Noisy input data from different settings, such as laboratories versus greenhouses, often leads to inaccuracies in shape estimation due to these domain shifts.

Various methods have been proposed to tackle these challenges. For instance, the CoRe method~\cite{magistri2022contrastive} employs a contrastive 3D shape completion technique that initially learns to generate the sweet pepper shape from a latent space. While effective in laboratory settings, it performs poorly in greenhouses. The HoMa~\cite{pan2023panoptic} framework, which generates both 3D shapes and fruit poses, offers better robustness against the irregular inputs typical of greenhouses. Another approach, T-CoRe~\cite{magistri2024efficient}, uses template matching to maintain fidelity to the typical sweet pepper shape, yet it struggles to predict accurate fruit geometry.

In this paper, we introduce CF-PRNet, a coarse-to-fine prototype refining network for point cloud completion and reconstruction. Our method innovatively applies a coarse-to-fine construction strategy, enhancing the model's ability to detail the sweet pepper shape progressively. We also implement a novel random input sampling strategy that selects a diverse array of frame observations to form the input point clouds. This strategy prevents model overfitting to limited variations of incomplete inputs and enhances generalization across different environmental conditions. Our experimental results underscore the effectiveness of CF-PRNet in addressing the challenges of data scarcity and domain shifts, significantly advancing the capabilities of AI in precision agriculture.

\begin{figure*}[t]
    \centering
    \includegraphics[width=0.99\linewidth]{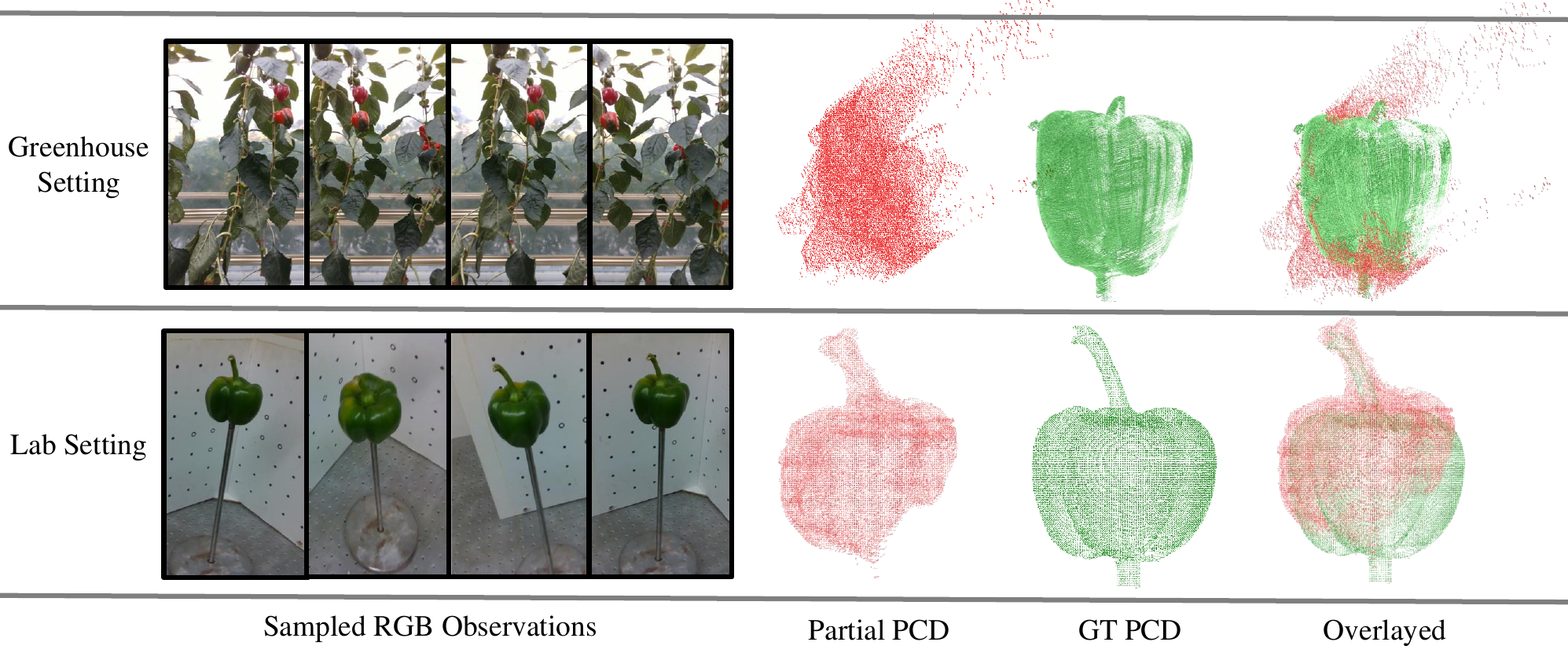}
    \vspace{-5pt}
    \caption{An illustration of the differences between greenhouse and laboratory settings. It can be seen that partial point clouds in greenhouse setting significantly diverge from the ground-truth point clouds. They could not provide similar supervision as the lab setting to the shape completion process, which pose a significant challenge for generalizing the model trained on laboratory data.}
    \label{fig:data}
    \vspace{-10pt}
\end{figure*}

\begin{figure*}[t]
    \centering
    \includegraphics[width=0.99\linewidth]{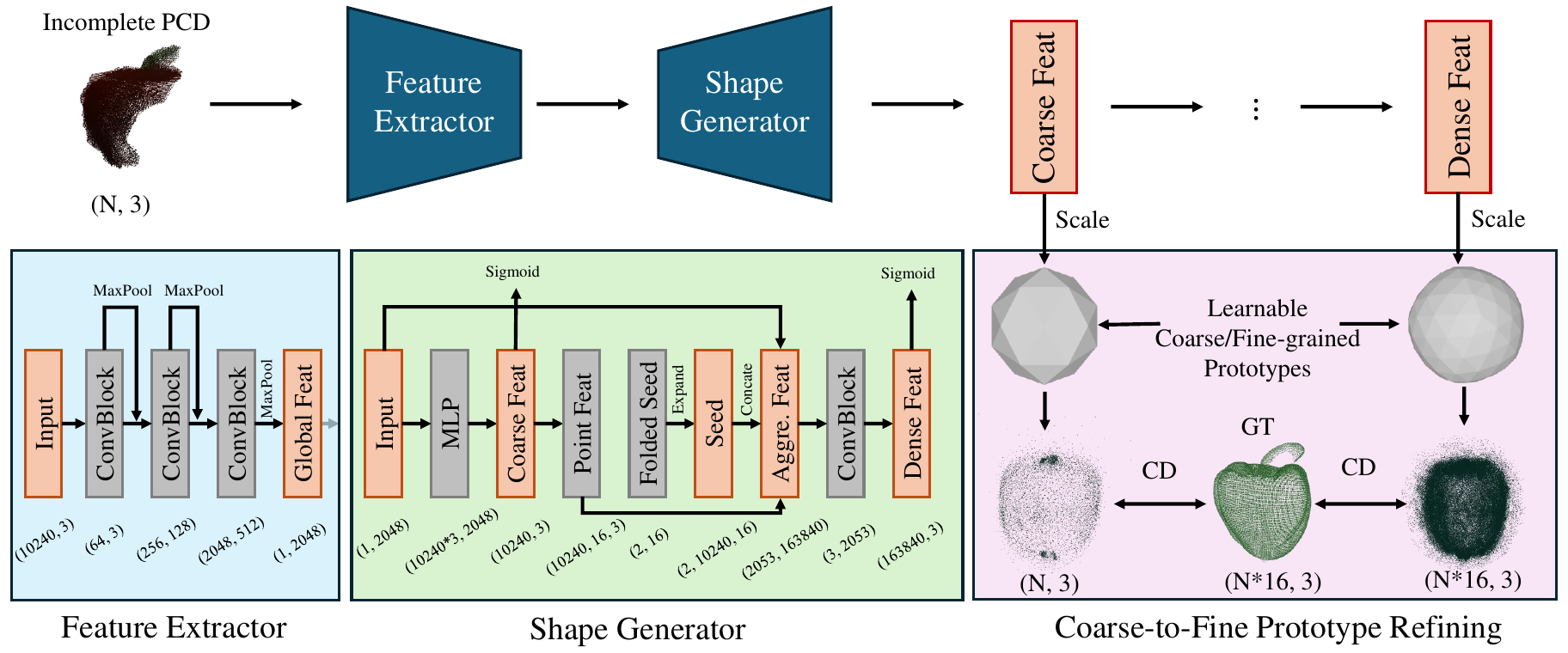}
    \caption{An illustration of CF-PRNet. The process begins with incomplete point clouds of sweet peppers fed into the feature extractor to obtain global features. These features are then processed by the shape generator to create coarse and dense modifications for the prototypes. The refining module fine-tunes these prototypes, progressively enhancing them from basic to detailed representations.}
    \label{fig:architecture}
    \vspace{-10pt}
\end{figure*}

\section{Methodology} 
\vspace{-10pt}
\noindent\textbf{Problem Definition.}
Let $\mS = (\vx, \vm, \vo )$ be a set of descriptors for an observation of a sweet pepper, including an RGB-D image $\vx \in \mathbb{R}^{H\times W\times 4}$, a binary mask map $\vm \in \mathbb{R}^{H\times W}$ and the camera pose information $\vo \in \mathbb{R}^{4\times 4}$. For each sweet pepper, there are a series of observations $\tS = \{\mS_1, \mS_2,\ldots,\mS_n ;\alpha\}$, where  $\alpha$ is the camera intrinsic parameters. With one of multiple sparse observations from $\tS$, we can leverage Open3D \cite{zhou2018open3d} to construct partial 3D point clouds $\mP$. When the observations are dense enough, we assume the ground-truth point clouds $\mY_g$ can be constructed.

In controlled laboratory settings, as depicted in \ref{fig:data}, constructing both partial and ground-truth point clouds (PCDs) to train shape completion models is straightforward. However, greenhouse settings often yield partial observations from limited viewing angles, resulting in noisy point clouds with reduced correlation to the ground-truth PCDs. This paper aims to bridge the discrepancy between laboratory and greenhouse settings by developing a shape completion network that generalizes effectively to the less controlled, more variable conditions of greenhouse environments.



\noindent\textbf{Coarse-to-Fine Point Completion Network.}
Our approach extends the point completion network framework~\cite{yuan2018pcn}, integrating three main modules: a feature extractor $f(\cdot)$, a shape generator $g(\cdot)$, and a novel coarse-to-fine prototype refining module $r(\cdot)$.


\noindent\textbf{Point Feature Extraction.}
The feature extractor $f(\cdot)$ processes partial 3D point clouds $\mP \in \mathbb{R}^{N \times 3}$ to derive a compact global feature vector $\vv = f(\mP) \in \mathbb{R}^{2048}$. This module consists of three stacked convolutional blocks, each equipped with 1D convolutional layers followed by batch normalization, ReLU activation, and another 1D convolutional layer. Max pooling is applied to the outputs of the intermediate layers, which are then concatenated with the subsequent layer inputs. The global features $\vv$, encapsulating the geometric information of the input point clouds, are ultimately acquired through max pooling at the final convolutional block's output.

\noindent\textbf{Shape Generator.}
The shape generator $g(\cdot)$ processes the global features $\vv$ from the point feature extractor, to produce coarse $\vc$ and dense $\vd$ features, denoted as $(\vc, \vd) = g(\vv)$. The initial stage of the shape generator incorporates an MLP block composed of three linear layers, interspersed with two ReLU activation functions. This MLP block expands the compact global features into coarse features $\vc$. These coarse features are then refined into high-resolution point features, aligning in dimensionality with the dense features $\vd$. Following~\cite{yuan2018pcn}, a `folded seed' is constructed to embed a generic prior about the sweet pepper's shape. This seed is expanded and merged with both the global and point features. The combined features are then processed through the final stage, a convolutional block consisting of three 1D convolutional layers, each followed by batch normalization and ReLU activation. The resulting output, the dense features $\vd$, are utilized to precisely scale the fine-grained prototype.

\noindent\textbf{Coarse-to-Fine Prototype Refining Module.}
The refining module employs coarse and dense features to adjust learnable sweet pepper prototypes. Initially, we generate two prototypes from 3D triangle meshes based on an icosahedral shape with a specific radius. These meshes undergo surface subdivision with varying iteration levels to form a coarse-grained prototype $\mT_{c}$ with 10,240 vertices and a fine-grained prototype $\mT_{d}$ with 163,840 vertices. The vertices serve as trainable parameters, while the mesh surfaces are retained for subsequent processing. To ensure the features are appropriately scaled, they are gated through a Sigmoid function. This gating mechanism adjusts the features to a suitable range, facilitating effective scaling of the prototypes. The final shapes, $\mY_c$ for the coarse and $mY_d$ for the dense features, are derived by performing an element-wise multiplication of the gated features with the prototype vertices. 

\noindent\textbf{Training Strategy}
To enhance robustness against noisy inputs typical of greenhouse environments, we adopt a distinct mapping strategy. Rather than converting all observations $\mathcal{T}$ of a sweet pepper into a unified point cloud, we map each individual observation $\mathcal{S}$ to its own point cloud object. During training, these point cloud objects are randomly combined in varying numbers to form a single training input. This method effectively increases the model's adaptability to the varied and unpredictable conditions found in greenhouse settings.

\noindent\textbf{Optimization}
There are three loss functions involved in training CF-PRNet, including Chamfer distance, normal consistency, and Laplacian smoothing. The Chamfer distance is applied for both coarse and fine-grained point clouds prediction, and are defined as:
\begin{equation}
\begin{gathered}
\mathcal{L}_{coarse} = \frac{1}{\lVert \mY_c \rVert} \sum_{\vp_c \in \mY_c}  \underset{\vp_g \in \mY_g}{\text{min}} \lVert \vp_c - \vp_g \rVert_2 + \frac{1}{\lVert \mY_g \rVert} \sum_{\vp_g \in \mY_g} \underset{\vp_c \in \mY_c}{\text{min}} \lVert \vp_g - \vp_c \rVert_2, \\
\mathcal{L}_{fine} = \frac{1}{\lVert \mY_d \rVert} \sum_{\vp_d \in \mY_d}  \underset{\vp_g \in \mY_g}{\text{min}} \lVert \vp_d - \vp_g \rVert_2 + \frac{1}{\lVert \mY_g \rVert} \sum_{\vp_g \in \mY_g} \underset{\vp_d \in \mY_d}{\text{min}} \lVert \vp_g - \vp_d \rVert_2,
\label{adaptive_attention}
\end{gathered}
\end{equation}
where $\vp \in \mathbb{R}^3$ represents a single point cloud.
With the modified prototypes as point cloud predictions, we assume the prototype surfaces connections between modified vertex remain steady. In this case, we can easily reconstruct a mesh with the point cloud predictions as new vertices and original surface information. To ensure smooth prediction, we enforce standard normal consistency and Laplacian smoothing losses on the meshes:
\begin{equation}
\begin{gathered}
\mathcal{L}_{norm} =  \sum_{i,j~\text{are adjacent}} (1 - \vn_i \cdot \vn_j)^2, 
\mathcal{L}_{lap} = \sum_{\vp_i \in \mY_d} \bigg\lVert \sum_{\vp_j \in \mN_i} \frac{1}{\lVert\mN_i \rVert} (\vp_i - \vp_j) \bigg\rVert_2, 
\end{gathered}
\end{equation}
where the normals $\vn_i$ and $\vn_j$ are associated with triangle faces. and $\mN_i$ is the neighboring point set of $\vp_i$. Overall, the training objective is 
\begin{equation}
\begin{gathered}
\mathcal{L}_{overall} = \lambda_1 \mathcal{L}_{coarse} + \lambda_2  \mathcal{L}_{fine} + \lambda_3  \mathcal{L}_{norm} + \lambda_4  \mathcal{L}_{lap},
\end{gathered}
\end{equation}
where $\lambda_1$,$\lambda_2$,$\lambda_3$,$\lambda_4$ are the coefficients of different loss functions.

\begin {table}[t]
\addtolength{\tabcolsep}{8pt}   
\caption {Sweet Pepper Completion results in the greenhouse setting. The $\uparrow$ and $\downarrow$ indicate that lower or higher values mean better performance.}
\vspace{-20pt}
\begin{center}
\scalebox{0.7}{
\begin{tabular}[t]{l r c c c c c}
\toprule
\multirow{2}*{\textbf{Methods}} & \multirow{2}*{\textbf{Venue}}   & \boldmath{$D_c$}[mm] & \textbf{F-score}[\%] &  \textbf{Precision}[\%] & \textbf{Recall}[\%]  & \multirow{2}*{\textbf{Learning?}}  \\
& & $\downarrow$avg & $\uparrow$avg & $\uparrow$avg  & $\uparrow$avg &    \\
\midrule
 CPD     \cite{myronenko2010point} & TPAMI'10      & 25.38 & 3.09  & 8.10  & 1.92  & \ding{55}  \\ 
 PF-SGD  \cite{marks2022precise}   & ICRA'22    & 9.28  & 35.03 & 37.32 & 33.21 & \ding{55}  \\ \midrule
 DeepSDF \cite{park2019deepsdf}    & CVPR'19    & 9.33  & 35.24 & 32.38 & 38.77 & \ding{51}  \\
 CoRe    \cite{magistri2022contrastive} & RA-L'22 & 6.90  & 41.47 & 43.17 & 41.64 & \ding{51}  \\
 HoMa    \cite{pan2023panoptic}         & IROS'23 & 5.29  & 58.56 & \textbf{61.28} & 56.26 & \ding{51}  \\
 T-CoRe  \cite{magistri2024efficient}   & ICRA'24 & 5.17  & 56.72 & 58.19 & 55.64 & \ding{51}  \\ \midrule
 CF-PRNet (ours)                        & CVPPA'24 & \textbf{3.78}  & \textbf{66.76} & {56.56} & \textbf{85.31} & \ding{51}  \\
\noalign{\smallskip}
\hline\bottomrule
\end{tabular}}
\end{center}
\label{main_performance}
\vspace{-20pt}
\end {table} 

\section{Experiments}
\noindent\textbf{Dataset.}
We conduct experiments on the sweet pepper benchmark dataset \cite{magistri2024dataset}. It consists of 129 different sweet peppers, of which 66 are used for training, and 25, 38 are used for validation and testing. The training set involves laboratory sweet peppers only, and the test set a from greenhouse only. The validation has a combination of lab and greenhouse sweet peppers, 16 and 9 respectively. The entire observation/frame numbers are 4580, 1387 and 980 for training, validation, and test set respectively. 


\vspace{5pt}
\noindent\textbf{Evaluation Metrics.}
Consistent with related work~\cite{magistri2022contrastive,pan2023panoptic}, we employ the Chamfer distance $D_c$, defined as the average symmetric squared distance between each point and its nearest neighbor in the opposing point cloud, to evaluate our shape completion solution. Additionally, F-score, precision, and recall are used at a fixed threshold for comprehensive quantitative assessment.

\vspace{5pt}
\noindent\textbf{Comparison with State-of-the-Art Methods.} We compare our method with existing methods as shown in Table \ref{main_performance}. In the four evaluation metrics, CF-PRNet outperforms the compared methods to a large margin. Particularly, we achieve over 10\% improvement on F-score over the second-best method T-CoRe~\cite{magistri2024efficient}. This performance boost is attributed to the significant improvement in recall, ~30\% improvement on the second best. We argue that this improvement is because our method is robust to the noisy input, and we are able to cover the entire shape of the complete sweet pepper. Although we sacrifice minor precision, the F-score is greatly improved. 

\vspace{5pt}
\noindent\textbf{Ablation Study.} As the test set of sweet pepper dataset is not publicly available after the challenge. We show the ablation results on the validation set only. To demonstrate the effectiveness of each component, we only choose the validation data from the greenhouse setting, which aligns better with the test set. CF-PRNet w/o Prototypes means we only use the coarse and fine-grained features to predict the complete shape. CF-PRNet w/o Coarse-to-Fine represents the variant without $\mathcal{L}_{coarse}$. CF-PRNet w/o Partial Sampling means we use all the observations to construct a point cloud input. The performance results demonstrate the effectiveness of the all the component in CF-PRNet.

\begin{figure*}[t]
    \centering
    \includegraphics[width=0.99\linewidth]{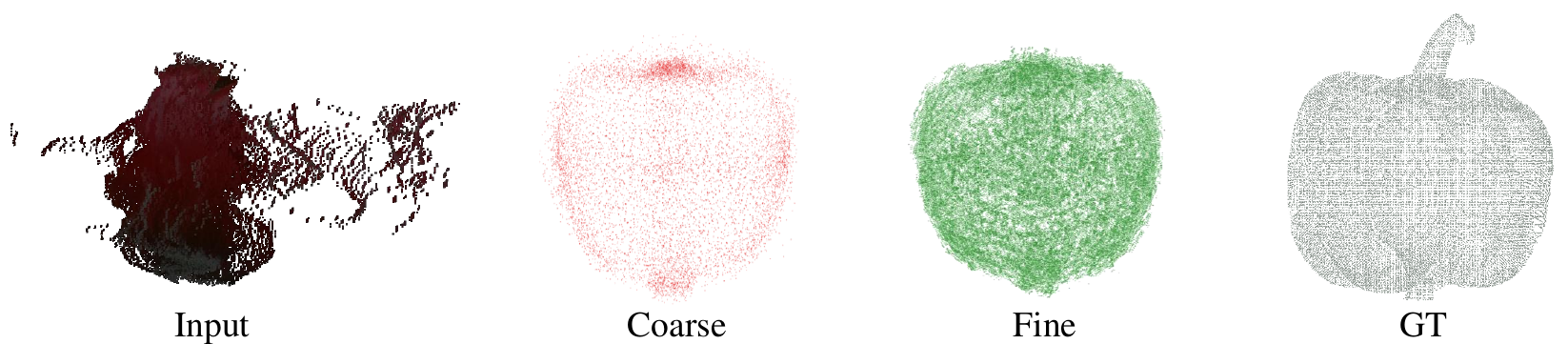}
    \caption{A visualization of the input, coarse and fine-grained output and the GT sample.}
    \label{fig:vis}
    \vspace{-10pt}
\end{figure*}

\vspace{5pt}
\noindent\textbf{Visualization.} In Fig. \ref{fig:vis}, we visualize the input and GT point clouds in the validation set, together with the coarse and fine-grained outputs from our model. It can be seen that the input diverges from the GT sample, but with the help of the prototypes, our output point clouds are still consistent with the output shape and size.

\begin {table}[t]
\addtolength{\tabcolsep}{8pt}   
\caption {Ablation Study of CF-PRNet on the validation set .}
\vspace{-20pt}
\begin{center}
\scalebox{0.7}{
\begin{tabular}[t]{l c c c c}
\toprule
\multirow{2}*{\textbf{Methods}} &  \boldmath{$D_c$}[mm] & \textbf{F-score}[\%] &  \textbf{Precision}[\%] & \textbf{Recall}[\%]   \\
&  $\downarrow$avg & $\uparrow$avg & $\uparrow$avg  & $\uparrow$avg     \\
\midrule
 CF-PRNet w/o Prototypes                       & \textbf{10.21}  & \textbf{29.24} & \textbf{24.98} & \textbf{35.26}  \\
 CF-PRNet w/o Coarse-to-Fine                   & \textbf{2.91}   & \textbf{74.08} & \textbf{64.02} & \textbf{92.31}  \\
 CF-PRNet w/o Partial Sampling                 & \textbf{3.06}   & \textbf{72.94} & \textbf{62.41} & \textbf{93.36}  \\
 CF-PRNet                                      & \textbf{2.59}   & \textbf{77.48} & \textbf{67.56} & \textbf{95.20}  \\
\noalign{\smallskip}
\hline\bottomrule
\end{tabular}}
\end{center}
\label{ablation}
\vspace{-20pt}
\end {table}

\vspace{-5pt}
\section{Conclusion}
In this study, we introduced CF-PRNet, a novel coarse-to-fine prototype refining network designed to address the challenging task of 3D shape completion for sweet peppers under partial observation scenarios, particularly in uncontrolled, greenhouse environments. Our approach innovatively combines the robustness of deep learning with the precision of traditional geometric methods through a dual-stage refinement process that utilizes both coarse and fine-grained prototypes.

%
%

\bibliographystyle{splncs04}
\bibliography{egbib}

\end{document}